\documentclass[10pt,twocolumn,letterpaper]{article}

\usepackage{cvpr}
\usepackage{times}
\usepackage{epsfig}
\usepackage{graphicx}
\usepackage{amsmath}
\usepackage{amssymb}
\usepackage{bbm}

\usepackage[dvipsnames]{xcolor}
\usepackage{mathtools}
\usepackage{multicol}
\usepackage{subcaption}
\usepackage{wrapfig}

\usepackage{booktabs}

\usepackage[pagebackref=true,breaklinks=true,letterpaper=true,colorlinks,bookmarks=false]{hyperref}

\cvprfinalcopy 

\def\cvprPaperID{6056}

\ifcvprfinal\fi

\begin{document}

\title{A Variational-Sequential Graph Autoencoder for Neural Architecture Performance Prediction}

\author{David Friede  \quad  Jovita Lukasik  \quad  Heiner Stuckenschmidt \quad Margret Keuper \\
\\
University of Mannheim}


\maketitle

\begin{abstract}
In computer vision research, the process of automating architecture engineering, Neural Architecture Search (NAS), has gained substantial interest. In the past, NAS was hardly accessible to researchers without access to large-scale compute systems, due to very long compute times for the recurrent search and evaluation of new candidate architectures. The NAS-Bench-101 dataset facilitates a paradigm change towards classical methods such as supervised learning to evaluate neural architectures. In this paper, we propose a graph encoder built upon Graph Neural Networks (GNN). We demonstrate the effectiveness of the proposed encoder on NAS performance prediction for seen architecture types as well an unseen ones (i.e., zero shot prediction). We also provide a new variational-sequential graph autoencoder (VS-GAE) based on the proposed graph encoder. The VS-GAE is specialized on encoding and decoding graphs of varying length utilizing GNNs. Experiments on different sampling methods show that the embedding space learned by our VS-GAE increases the stability on the accuracy prediction task. 
\end{abstract}

\section{Introduction}

Deep learning using convolutional neural architectures has been the driving force of recent progress in computer vision and related domains. Multiple interdependent aspects such as the increasing availability of training data and compute resources are responsible for this success. Arguably, none has had as much impact as the advancement of novel neural architectures \cite{krizhevsky2012imagenet,goodfellow2014generative}. Thus, the focus of computer vision research has shifted from a feature engineering process to an architecture engineering process. The urge to automate this process using machine learning techniques is a direct consequence.
\textit{Neural Architecture Search} (NAS) \cite{elsken2018neural} attends to techniques automating architecture engineering. Due to very long compute times for the recurrent search and evaluation of new candidate architectures for example in proposed genetic algorithms or reinforcement learning~\cite{zoph2018learning}
, NAS research has hardly been accessible for researchers without access to large-scale compute systems.

Yet, the publication of \textit{NAS-Bench-101} \cite{ying2019bench}, a dataset of over $423$k fully trained neural architectures, facilitates a paradigm change in NAS research. Instead of carefully evaluating each new proposed neural architecture, NAS-Bench-101 enables to experiment with classical data-based methods such as supervised learning to evaluate neural architectures. In this paper, we tackle the task of learning to predict the accuracy of convolutional neural architectures in a supervised way using continuous representations of neural architectures. We further demonstrate that such embeddings can be employed for the reconstruction and generation of neural architectures. In combination, these two abilities allow to sample neural architectures at a given target accuracy.

Most current neural architectures for computer vision can be represented as directed, acyclic graphs (DAGs). Thus, we base our approach on Graph Neural Networks. \textit{Graph Neural Networks} (GNNs) \cite{wu2019comprehensive} have proven very powerful comprehending local node features and graph substructures. This makes them a very useful tool to embed nodes as well as full graphs like the NAS-Bench-101 architectures into continuous spaces.


Within this setting, finding new neural architectures translates into generating new graphs. From the few existing graph generating models, sequential approaches like \cite{you2018graphrnn} or \cite{li2018learning} are very promising. The model \textit{Deep Generative Models of Graphs} (DGMG) in \cite{li2018learning} utilizes GNNs and shows dominance over Recurrent Neural Network (RNN) methods. In natural language processing, it is common to use RNNs on an encoder-level as well as on a decoder-level. Inspired by this approach, we extend the ideas of the DGMG model to build a \textit{Variational-Sequential Graph Autoencoder} (VS-GAE), a variational autoencoder \cite{kingma2013auto} that utilizes GNNs on the encoder-level and decoder-level simultaneously. To the best of our knowledge, we propose the first graph autoencoder that is built on GNNs and acts on graphs of different sizes. This makes it a powerful tool to handle neural architectures.

Having a densely sampled and fully evaluated search space at hand, such as provided by NAS-Bench-101, is a luxury that is rarely given for real-world problems. Thus, sampling methods play an important role when working on neural architectures. While current sampling driven approaches to NAS~ \cite{Guo2019} rely on uniform or equidistant sampling in the discrete search space defined through the edit distance, we argue that such a space is artificial by nature and does not reflect the similarities between architectures appropriately. If for example a $3\times 3$ convolution is replaced by a $5\times 5$ convolution, the distance is the same as if it were replaced by a pooling operation. We thus propose to sample unseen neural architectures from our learned VS-GAE \footnote{In this scenario, VS-GAE is obviously trained without supervision.} embedding space as an alternative and show increased stability on the accuracy prediction task.  
In summary, we make the following contributions:
\begin{itemize}
\item We present a graph encoder built on GNNs and adjusted to the NAS-Bench-101 neural architectures. We outline the benefit of this graph encoder for performance prediction on the one hand and for modeling a graph decoder on the other hand.
\item We provide VS-GAE, a new variational-sequential graph autoencoder. VS-GAE is specialized on the encoding and decoding of graphs of varying length utilizing GNNs. This makes it a strong tool for designing new architectures and generating a meaningful graph latent space.
\item We outline different sampling methods. Using VS-GAE, we introduce a method to sample in the graph latent space. We compare this new sampling approach with existing methods based on the discrete graph space and show increased stability.
\end{itemize}
The remaining paper is structured as follows: Section \ref{sec:related} gives a short review of the related work. In Section \ref{sec:between}, we further discuss GNNs and graph generating models. In Sections \ref{sec:encoder} - \ref{sec:sampling}, we present our proposed model. We introduce the encoder, our model VS-GAE and the sampling methods. In Section \ref{sec:experiments}, we present our experiments and results. Finally, we give a conclusion and outline some future directions in Section \ref{sec:conclusion}.

\section{Related Work} \label{sec:related}
Neural Architecture Search (NAS) \cite{Zoph2017, Real2017, zoph2018learning, Liu2018, Pham2018}, the process of designing neural network architectures in an automatic way, gained substantial attention recently. See \cite{elsken2018neural} for an overview and detailed survey over recent NAS methods. The currently most successful approaches follow different paradigms: Reinforcement learning (RL) \cite{Zoph2017, zoph2018learning,Pham2018} as a NAS strategy considers the neural architecture generation as the agent's action with it's reward given in terms of validation accuracy. Evolutionary Algorithm (EA) \cite{Real2017, Liu2017} approaches optimize the neural architectures themselves by guiding the mutation of architectures and evaluating their fitness given in terms of validation accuracy. Bayesian optimization (BO) \cite{Kandasamy2018} derive kernels for architecture similarity measurements to extrapolate the search space. Gradient based methods \cite{Liu2018, luo2018neural} use continuous relaxations of neural architectures to allow for gradient-based optimization.

NAS-Bench-101 \cite{ying2019bench} is a public dataset of neural architectures in a restricted cell structured search space \cite{zoph2018learning} evaluated on the CIFAR-10-classification set \cite{Krizhevsky2012}. NAS-Bench-101 considers the following constraints to limit the search space: it only considers directed acyclic graphs, the number of nodes is limited to $ \vert V\vert \leq 7$, the number of edges is limited to $\vert E\vert \leq 9$ and only 3 different operations are allowed $\{3~\times~3 ~\mathrm{convolution}, 1~\times~1 ~\mathrm{convolution}, 3~\times~3 ~\mathrm{max-pool} \}$. These restrictions lead to a total of $423$k  unique convolutional architectures. The architectures have been trained for four increasing numbers of epochs $\{4,12,36,108\}$. Each of these architectures is mapped to its validation and training measures.  In this paper we use the architectures trained for $108$ epochs and their corresponding validation accuracy.

The idea of Graph Neural Networks as an iterative process which propagates the node states until an equilibrium is reached,  was initially mentioned 2005 in \cite{gori2005new}. Years later when CNNs became popular, \cite{bruna2013spectral} and \cite{henaff2015deep} defined graph convolutions in the Fourier domain by utilizing the graph Laplacian. The modern interpretation of GNNs was first mentioned in \cite{niepert2016learning} and \cite{kipf2016semi} where node information was inductively updated through aggregating information of each node's neighborhood. This approach was further specified and generalized by \cite{hamilton2017inductive} and \cite{gilmer2017neural}.

Existing graph generating models can roughly be classified in global approaches and sequential approaches. Global approaches output the full graph at once usually by relaxing the adjacency matrix \cite{kipf2016variational, simonovsky2018graphvae}. The sequential approach is an iterative process of adding nodes and edges alternately. \cite{luo2018neural} used RNNs to generate neural architectures in this sequential manner. The model in \cite{you2018graphrnn} introduced a second edge-level RNN capturing the edge dependencies. \cite{zhang2019d} created and used an asynchronous message passing scheme instead of RNNs to decode the computations of neural architectures. In contrast, the model in \cite{li2018learning} utilizes the synchronous message passing scheme as known from GNNs for the sequential graph generation.

Variational autoencoders (VAE) \cite{kingma2013auto} allow to learn a recognition model $q_{\phi}(\textbf{z} \vert G)$, i.e., the encoder, and a generative model $p_{\theta}(G \vert \textbf{z})$, i.e., the decoder, jointly, where $\phi$ and $\theta$ are the outputs of the encoder and decoder, respectively. Note, the VAE learns the parameters of the data's probability distribution instead of a lower dimensional representation of the input data.

The VAE learns by maximizing a lower bound estimator on the log-likelihood $\text{log } p_{\theta}(G)$:
\begin{eqnarray}\label{VAE}
{\mathcal{L}}(\theta, \phi ; G) &=& \mathbb{E}_{q_{\phi}(\textbf{z} \vert  G)} \big[\log p_{\theta}(G \vert \textbf{z}) \big]  \nonumber \\
&-& \text{D}_{\text{KL}} (q_{\phi}(\textbf{z} \vert G) \Vert p_{\theta}(\textbf{z}))
\end{eqnarray}	
From the trained VAE model new data can be generated by sampling $\textbf{z} \sim p_{\theta}(\textbf{z})$ and decode it by means of $p_{\theta}(G \vert \textbf{z})$. 

\section{Continuous Graph Representations}\label{sec:between}
Combining modern machine learning methods with graph structured data has increasingly gaining popularity. And rightly so; one can interpret it as an extension of deep learning techniques to non-Euclidean data \cite{bronstein2017geometric} or even as inducing relational biases within deep learning architectures to enable combinatorial generalization \cite{battaglia2018relational}. Because of the discrete nature of graphs, they can not trivially be optimized in differentiable learning methods that act on continuous spaces. In this paper, we address this problem. We want to use continuous methods to handle the graphs characterizing neural architectures from the NAS-Bench-101 dataset.

\subsection{From Discrete to Continuous}
The research in GNNs enabled breakthroughs in multiple areas related to graph analysis such as computer vision \cite{xu2017scene, landrieu2018large,yi2017syncspeccnn}, natural language processing \cite{bastings2017graph}, recommender systems \cite{monti2017geometric}, chemistry \cite{gilmer2017neural} and others. The capability of GNNs to accurately model dependencies between nodes makes them the foundation of our research. We utilize them to move from the discrete graph space to the continuous space and vice versa.

\subsection{From Continuous to Discrete}
Generating new graphs particularly new neural architectures is an ambitious task that has to overcome multiple fundamental challenges. The main focus lies in the highly variable graph search space of NAS-Bench-101 and the complex dependencies within a single graph.

Global approaches like \cite{kipf2016variational} or \cite{simonovsky2018graphvae} are restricted to a fixed and small number of nodes since they utilize relaxations of the adjacency matrix which is by nature quadratic in the number of nodes. \cite{luo2018neural} were the first to use RNNs to generate neural architectures in a sequential manner. Their model acts on graphs with a fixed number of nodes and lacks the capability to induce the complex graph structures. These issues were partially addressed in \cite{you2018graphrnn}. This model acts on graphs of variable sizes and utilizes a second edge-level RNN to capture the edge dependencies. In \cite{li2018learning}, the superiority of using GNNs over RNNs during the graph generating process was expressed.

Our model can be interpreted as an extension of the conditional version of \cite{li2018learning}. To the best of our knowledge, our model is the first that utilizes GNNs on the encoding-level as well as the decoding-level obtaining a variational-sequential graph autoencoder acting on graphs of different sizes. The model in \cite{zhang2019d} is also related to our work; unlike our model, it acts on a fixed number of nodes. Also in contrast to our method, \cite{zhang2019d} built a model on an asynchronous message passing scheme that encodes computations instead of graph structures.


\color{black}
\section{The Graph Encoder}\label{sec:encoder}
\begin{figure}[t]
\begin{center}
\includegraphics[width=\linewidth]{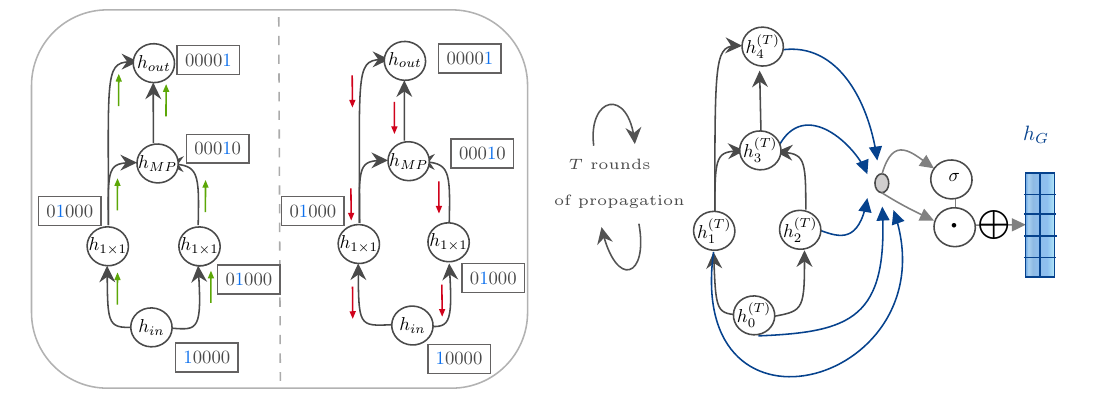}
\end{center}
   \caption{Illustration of the graph encoding process: The node-level propagation using $T$ rounds of bidirectional message passing (left) and the graph-level aggregation into a single graph embedding $h_G$ (right).}
\label{fig:encoder}
\end{figure}

In this section we present our GNN-based model to encode the discrete graph space of NAS-Bench-101 into a continuous vector space. One can imagine a single GNN iteration as a two-step procedure. First, each node sends out a message to its neighbors alongside its edges. Second, each node aggregates all incoming messages to update itself. After a final amount of these iteration steps, the individual node embeddings are aggregated into a single graph embedding.

\subsection{Node-Level Propagation} Let $G=(V,E)$ be a graph with nodes $v\in V$ and edges $e\in E\subseteq V\times V$. We denote $N(v)=\{u\in V\mid (u,v)\in E\}$ and $N^{out}(v)=\{u\in V\mid (v,u)\in E\}$ as the directed neighborhoods of a node $v \in V$. For each node $v\in V$, we associate an initial node embedding $h_v\in \mathbb{R}^{d_n}$. In our experiments we use a learnable look-up table based on the node types. Propagating information through the graph can be seen as an iterative \textit{message-passing} process
\begin{align}
m_{u\rightarrow v} &=  \Xi_{u\in N(v)}\bigl(M^{(t)}(h_v^{(t-1)}, h_u^{(t-1)})\bigr),\label{M module} \\ 
h_v^{(t)} &= U^{(t)}(h_v^{(t-1)}, m_{u\rightarrow v}), \label{U module}
\end{align}

with a differentiable message module $M^{(t)}$ in (\ref{M module}), a differentiable update module $U^{(t)}$ in (\ref{U module}) and a differentiable, permutation invariant aggregation function $\Xi$. The message module $M^{(t)}$ is illustrated by the green arrows in Figure \ref{fig:encoder} (left). To address the directed nature of the NAS-Bench-101 graphs, we add a reverse message module
\begin{align}
m^{out}_{u\rightarrow v} &= \Xi_{u\in N^{out}(v)}\bigl(M_{out}^{(t)}(h_v^{(t-1)}, h_u^{(t-1)}\bigr),\label{M rev}\\
h_v^{(t)} &= U^{(t)}(h_v^{(t-1)}, m_{u\rightarrow v}, m^{out}_{u\rightarrow v}).
\end{align}
This is outlined in Figure \ref{fig:encoder} (left) by the red arrows and leads to so-called bidirectional
message passing. The update module $U^{(t)}$ utilizes each node's incoming messages to update that node's embedding from $h_v^{(t-1)}$ to $h_v^{(t)}$.

Exploring many different choices for the message and update modules experimentally, we find that the settings similar to \cite{li2018learning} work best for our needs. We pick a concatenation together with a single linear layer for our message modules. The update module consists of a single gated recurrent unit (GRU) where $h_v^{(t-1)}$ is treated as the hidden state. For the aggregation function, we choose the sum. To increase the capacity of our model, on the one hand, we apply multiple rounds of propagation and on the other hand, we use a different set of parameters for each round.

\subsection{Graph-Level Aggregation} After the final round of message-passing, the propagated node embeddings $h=(h_v)_{v\in V}$ are aggregated into a single graph embedding $h_G\in \mathbb{R}^{d_g}$, where

\noindent\begin{minipage}{.5\linewidth}
  \begin{align}\label{graph aggr}
  	h_G = A(h),
  \end{align}
\end{minipage}%
\begin{minipage}{.5\linewidth}
  \begin{align}\label{graph aggr var}
  	h^{\mathrm{var}}_G = \Tilde{A}(h).
  \end{align}
\end{minipage}
\vskip\baselineskip
We obtain good results by using a linear layer combined with a gating layer that adjusts each node's fraction in the graph embedding. This aggregation layer $A$ in (\ref{graph aggr}) is further illustrated in Figure \ref{fig:encoder} (right). In case that variational outputs are required, we interpret $h_G$ as the mean and add an extra graph aggregation layer $\Tilde{A}$ in (\ref{graph aggr var}) which outputs the variance $h^\mathrm{var}_G$.

\section{VS-GAE}
\begin{figure*}[t]
\begin{center}
\includegraphics[width=\linewidth]{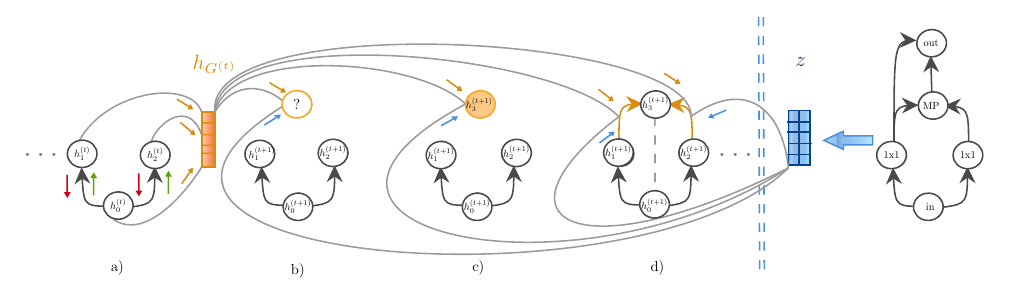}
\end{center}
   \caption{Illustration of a single iteration during the graph generation process. a) A decoder-level GNN propagates the node embeddings through the partially created graph and aggregates them into a summary of this graph. b) A new node is created and its node type is selected using the summary of the partially created graph as well as the original one. c) The newly created node is initialized with a node embedding. d) A score of all edges connecting the new node is calculated and evaluated into the set of new edges.}
\label{fig:decoder}
\end{figure*}

In this section we present our Variational-Sequential Graph Autoencoder (VS-GAE). The VS-GAE is composed of the variational version of the graph-based encoder from Section~\ref{sec:encoder} and a graph generating decoder using a sequential process. The encoder $q_{\phi}(\textbf{z} \vert G)$ takes graph $G$ with node labels as input and outputs a prior distribution $p(\textbf{z})$ over the latent space in line with common variational autoencoders. The decoder $p_{\theta}(G \vert \textbf{z})$ takes a sampled point $\textbf{z}$ from this latent space $p(\textbf{z})$ as input and generates a graph iteratively as a sequence of operations that add new nodes and edges until the end/output node is generated. Note that the sampled point $\textbf{z}$ contains a summary of the original graph $G$.

\subsection{Graph Generating Process}
The decoder consists of multiple modules, mainly describing a distribution over the outcomes of a specific step in the generating sequence. Each module utilizes for each iteration $t$ one or multiple of the following inputs:
\begin{itemize}
    \itemsep0em 
	\item[]{\makebox[2cm]{$\textbf{z}$\hfill} the sampled point from the latent space,}
	\item[]{\makebox[2cm]{$L$\hfill} a look-up table based on the node types,}
	\item[]{\makebox[2cm]{$h^{(t)}$\hfill} the embedding of the created nodes,}
	\item[]{\makebox[2cm]{$G^{(t)}, h_{G^{(t)}}$\hfill} the partial graph and its embedding.}
\end{itemize}
Note that the learnable embedding look-up table $L$ is independent of the one in Section~\ref{sec:encoder}. Correspondent to the NAS-Bench-101 graphs, we begin the iteration with the start/input node that receives an initial node embedding $h^{(0)}=(h_0)$ according to the sampled point and the look-up table. With these preparations we can represent the full graph generating process through iterating over the following modules. Such iteration step can be tracked module by module through following Figure~\ref{fig:decoder}. Note that the modules' weights are shared over different iterations.
\paragraph{GraphProp}
This module processes the embedding $h^{(t)}$ of the previously created nodes together with their underlying graph structure $G^{(t)}$ for two complementary but distinct tasks. First, the node embeddings are updated by propagating through them. Second, the updated node embeddings are read out and aggregated into a single graph embedding,
\begin{align}\label{mod:prop}
h_{G^{(t)}},h^{(t)}_p  = f_{\mathrm{prop}}(h^{(t)}, G^{(t)}).
\end{align}
This is illustrated in Figure~\ref{fig:decoder} a). The graph embedding $h_{G^{(t)}}$ can be interpreted as a summary of the hitherto created partial graph. We use a GNN to propagate and aggregate the node embeddings $h^{(t)}$ in (\ref{mod:prop}). More precisely, we use an exact copy of our encoder from Section~\ref{sec:encoder} initialized with its own weights. This is motivated by NLP methods that use two distinct RNNs on the encoder-level and the decoder-level, respectively.

\paragraph{AddNode}\label{AddNode}
In this module, a new node is created and its node type is selected. The input is the summary of the original graph, i.e., the sampled point $\textbf{z}$ from the latent space as well as the summary $h_{G^{(t)}}$ of the already created partial graph. The intention behind these inputs and the ones for the following modules is always the same: What does the original graph look like? What does the partially created graph look like? What is missing? This concludes in following module,
\begin{align}\label{NodeType}
\mathrm{NodeType} = \mathrm{softmax}\bigl(f_{\mathrm{addNode}}(\textbf{z}, h_{G^{(t)}})\bigr).
\end{align}
The \textit{addNode} module is outlined in Figure \ref{fig:decoder} b). The output of (\ref{NodeType}) is a categorical distribution over all possible node types. Note that sampling over this distribution yields a one-hot encoding of a specific node type. In line with the structure of NAS-Bench-101 graphs, the iteration stops after running through the step that adds the end/output node.

\paragraph{InitNode}
This module initializes the node embedding of the just created node. The input is the sampled point $\textbf{z}$, the summary $h_{G^{(t)}}$ of the created graph and the embedding of the node type $L[\mathrm{type}]$ in the look-up table. This embedding of the new node is then added to the already existing propagated node embeddings,
\begin{align}
h_{t+1} &= f_{\mathrm{initNode}}(\textbf{z}, h_{G^{(t)}}, L[\mathrm{type}]),\\
h^{(t+1)} &= (h^{(t)}_{p,1},\dots,h^{(t)}_{p,t}, h_{t+1}).
\end{align}
This process is further illustrated in Figure \ref{fig:decoder} c).

\paragraph{AddEdges} \label{module AddEdges}
In this module, the edges towards the newly created node are selected. For this purpose, a score between the new node $h_{t+1}$ and each previous node is calculated, respectively. A high score stands for a high probability of an edge and vice versa. This is illustrated in Figure~\ref{fig:decoder} d).
\begin{align}
s_v &= f_{\mathrm{addEdges}}(h_{t+1}, h_v, \textbf{z}, h_{G^{(t)}}),~ \ h_v\in h^{(t)}_p, \\
\mathrm{Edges} &= \sigma(s), \label{AddEdges}
\end{align}
where \textit{Edges} is a family of Bernoulli distributed random variables describing a probability for each possible edge $e$ connecting the new node. Sampling over these distributions yields the new set of edges. 
We interpret each edge as directed towards the new node.

In all our experiments, we let $f_{\mathrm{addNode}}, f_{\mathrm{initNode}}$ and $f_{\mathrm{addEdges}}$ be two-layer MLPs with ReLU nonlinearities.

\subsection{Training and Loss}
Recall that a VAE maximizes the lower bound estimator for a graph $G$ and a latent space representation $\textbf{z}$:
\begin{eqnarray} \label{VS-GAE-Loss}
{\mathcal{L}}(\theta, \phi ;G) &=& \mathbb{E}_{q_{\phi}(\textbf{z} \vert G)} \big[\log p_{\theta}(G \vert \textbf{z}) \big]  \nonumber \\
&+& \text{D}_{\text{KL}} (q_{\phi}(\textbf{z} \vert G) \Vert p_{\theta}(\textbf{z})).
\end{eqnarray}
The first term of (\ref{VS-GAE-Loss}) is the model specific reconstruction loss which enforces high similarity between the input graph and the generated graph. The second term is the Kullback--Leibler Divergence which regularizes the latent space. In the following, we will discuss the reconstruction loss of VS-GAE.


We train the encoder and the decoder of VS-GAE jointly in an unsupervised manner. Although the encoder is by construction invariant under graph isomorphisms, the decoder needs a certain ordering over the nodes. To fulfill the prior of the decoder that each edge is directed towards the new node, this ordering has to be in such a manner that the adjacency matrix is an upper triangle matrix. This is, for example, given by the canonical ordering in which the graphs of NAS-Bench-101 are provided.

Given this fixed ordering of the nodes, we know the ground truth of the outputs of \textit{AddNode} (\ref{NodeType}) and \textit{AddEdges} (\ref{AddEdges}) during training. One the one hand, we can use this ground truth to calculate a node-level loss $\mathcal{L}_V^i$ and an edge-level loss $\mathcal{L}_{E}^i$ at each iteration step, respectively. On the other hand, we can replace the model's output by the ground truth such that possible errors will not accumulate through iterations. This is also known as teacher forcing.

In order to calculate the overall reconstruction loss for a graph $G$, we sum up node losses and edge losses over all iterations
\begin{align}
\mathcal{L}_{rec} = \mathcal{L}_V + \mathcal{L}_{E}.
\end{align}

Following \cite{kingma2013auto}, we assume $p_{\theta}(\textbf{z})~\sim~\mathcal{N}(\textbf{z};0, \mathbbm{1})$ and $p_{\theta}(G \vert \textbf{z}) \sim \mathcal{N}(\mu, \Sigma)$. Furthermore, we approximate the posterior by a multivariate Gaussian distribution with diagonal covariance structure. This can be written as $\log q_{\phi}(\textbf{z} \vert G) = \log \mathcal{N}(\textbf{z};\mu, \sigma^{2} \mathbbm{1})$ and ensures a closed form of the  $\mathrm{KL}$ divergence
\begin{eqnarray} 
\text{D}_{\text{KL}}=- \frac{1}{2} \sum_{j=1}^{J} (1+ \log ((\sigma_j)^2) - (\mu_j)^2 - (\sigma_j)^2).
\end{eqnarray}
Thus, it holds for the overall loss
 \begin{eqnarray} 
{\mathcal{L}} = \mathcal{L}_V + \mathcal{L}_{E} + \alpha\text{D}_{\text{KL}},
\end{eqnarray}
where the $\mathrm{KL}$ divergence is additionally regularized. Following previous work of \cite{Jin2018} and \cite{zhang2019d}, we set the regularization term $\alpha=0.005$.

\section{Sampling Methods} \label{sec:sampling}
In this section we present multiple sampling methods for the NAS-Bench-101 dataset. We distinguish between methods acting directly in the discrete graph space and a method in the continuous latent space utilizing the VS-GAE embedding.
 
\subsection{Discrete Space}
A fundamental property of the NAS-Bench-101 dataset is the varying graph size from $2$ to $7$ nodes. We make use of this feature to implement a simple sampling method. For each set of graphs with the same node size we sample uniformly at random.

The second approach is a more sophisticated sampling method in the discrete space. A recent work \cite{Guo2019} finds that sampling uniformly with respect to the graph space constraints performs better than the purely random search. For this purpose, we make use of the edit distance. The edit distance is the smallest number of changes which are required to transform one graph into another; one change consists of either turning an operation, i.e., the node's label, or adding and removing an edge, respectively. Edit sampling consists of sampling uniformly by means of the edit distance of graphs with the same length.

\subsection{Continuous Space}
\begin{figure}
    \centering
    \begin{subfigure}[b]{0.23\textwidth}
        \includegraphics[width=\textwidth]{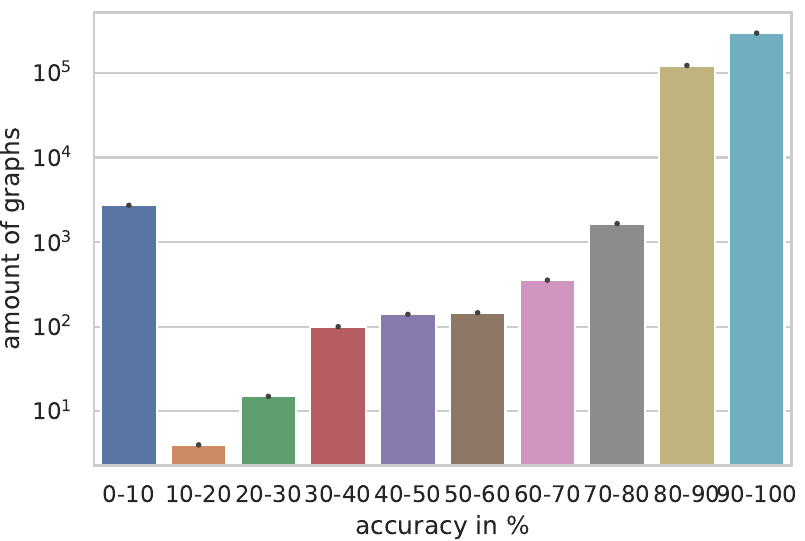}
    \end{subfigure}
    \begin{subfigure}[b]{0.23\textwidth}
        \includegraphics[width=\textwidth]{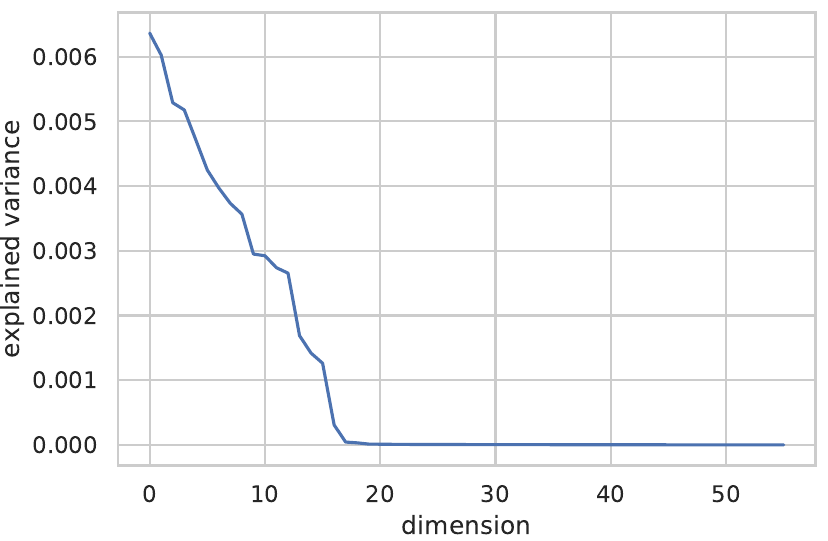}
    \end{subfigure}
    \caption{Two distinct properties of NAS-Bench-101; The allocation of the dataset sorted by the ground truth accuracy in logarithmic scale (left) $\sim$98.8\% in the two last bins. The explained variance of the VS-GAE latent space regarding a principal component analysis (right) $\sim$16 significant dimensions.}\label{fig:numbers}
\end{figure}

For the continuous approach, we fully train VS-GAE without supervision on the NAS-Bench-101 dataset and pick $d_G=56$ as the dimension of the latent space. Through its variational nature, the latent space builds upon a probability distribution space $p(\textbf{z})$. For our sampling method, we transform this distribution space into a regular vector space by discarding the variance, i.e., each graph is now represented by its mean derived by the encoder. The principal components of the embedded graphs in this vector space, see Figure \ref{fig:numbers} (right), suggest that it suffices to reduce the dimension drastically. By reducing the dimensions to $4$,  we keep enough information to define a useful sampling method while preventing the curse of dimensionality regarding equidistant sampling.

This sampling method operates as follows: We divide the reduced vector space into a fixed number of bins of same size.  The number of bins is detected in a preprocessing step, where we aim to have roughly the same amount of non-empty bins as required number of samples in order to reduce the compute time. To obtain randomness, we shift these bins by a random proportion in $[0,1]$ of the bin size in each dimension, respectively. For each non-empty bin, we select the graph whose embedding is the closest to the center of the bin.
In case of too many samples, we discard the redundant samples randomly. 
If not enough graphs are sampled, i.e., the number of non-empty bins is less than the required amount of samples, we have to differentiate between two scenarios: first, if the number of non-empty bins is greater than the number of missing samples $k$, we pick $k$ non-empty bins randomly and sample one graph of these bins randomly. Second, if the number of missing samples exceeds the number of non-empty bins, we randomly sample graphs from each of these bins and repeat the procedure until we arrive at the first scenario.



\section{Experiments}\label{sec:experiments}

Our experiments aim to cover three different but still complementary domains. The core of VS-GAE is its encoder since the full model is built around it. Our first experiment evaluates the encoder's performance prediction ability. For the second experiment, we test the abilities of VS-GAE itself. Last, we unite all discussed theories and show experimentally how sampling in the latent space generated by VS-GAE is a useful tool to reinforce the performance prediction of the encoder.

If not mentioned differently, we set $d_v=250$ for the node dimensions and $d_G=56$ for the dimension of the latent space. We split the dataset $70\%$/$20\%$/$10\%$ edit-sampled into training-, test- and validation set.

All our experiments are implemented using PyTorch \cite{paszke2017automatic} and PyTorch Geometric \cite{fey2019fast}. Our code is available at : \url{https://github.com/jovitalukasik/vs_gae}

\subsection{Performance Prediction}

\begin{figure}
  \centering
    \includegraphics[width=0.475\textwidth]{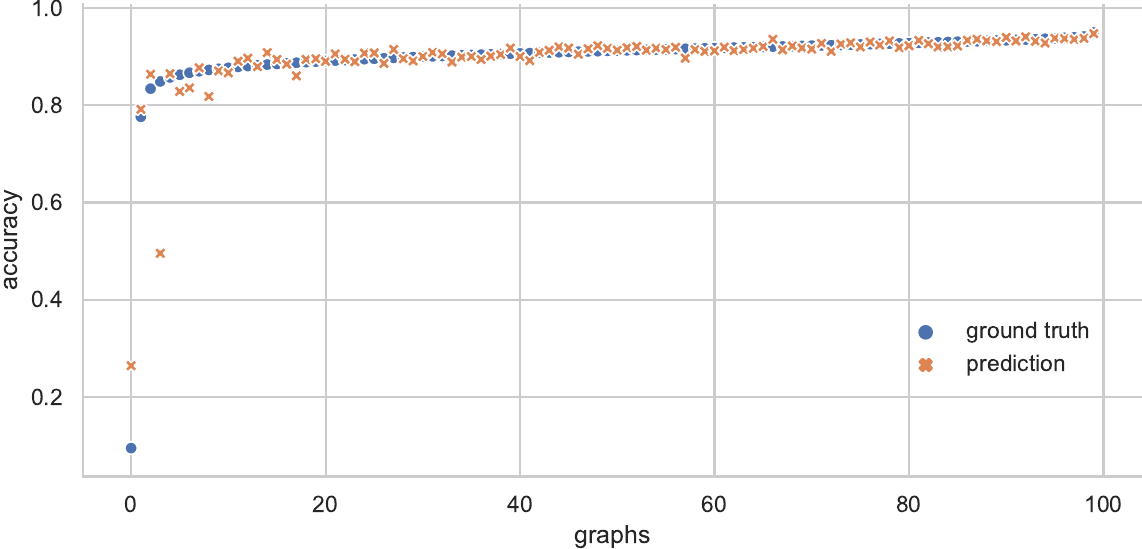}
  \caption{The predicted accuracy and ground truth of 100 randomly sampled graphs from the NAS-Bench-101 dataset sorted by the ground truth showing a low prediction error for graphs with high accuracy. For low accuracy architectures, our model mostly predicts low values.} \label{fig:graphs}
\end{figure}

\begin{figure}
  \centering
    \includegraphics[width=0.475\textwidth]{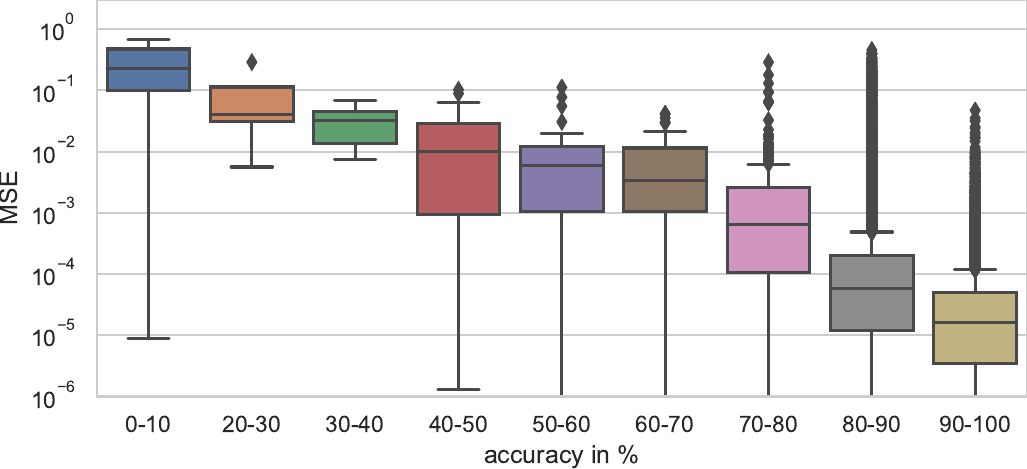}
  \caption{The mean and variance of the squared error of the test set performance prediction sorted by the ground truth accuracy in logarithmic scale. Predictions are less certain for architectures in the low accuracy domain.} \label{fig:losses}
\end{figure}

\begin{figure}
  \centering
    \includegraphics[width=0.475\textwidth]{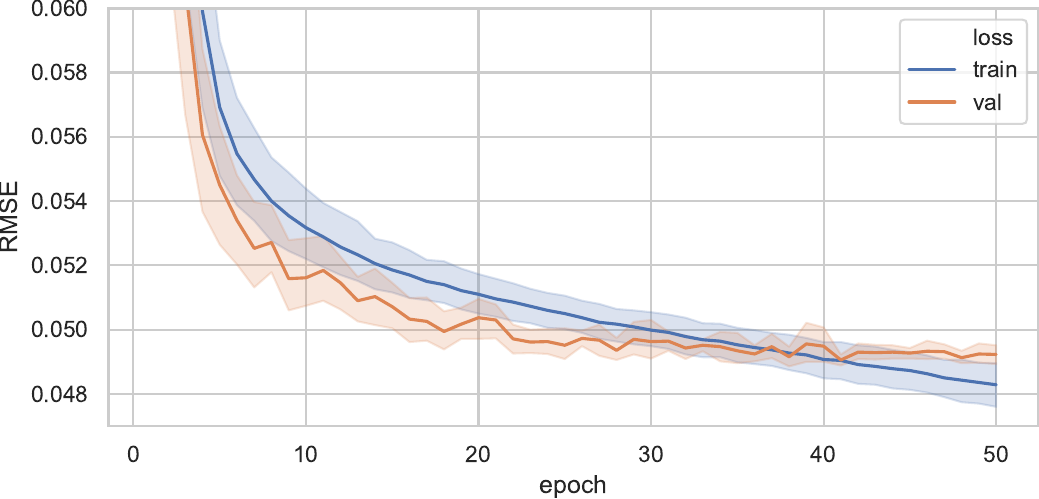} 
  \caption{Progress of loss and validation error over 50 epochs regarding performance prediction. Best validation RMSE $\sim0.0487$.} \label{fig:Interpolation}
\end{figure}

In these experiments, we evaluate how well the latent space generated by the encoder can predict a metric of interest of the NAS-Bench-101 graphs, i.e., the validation accuracy on the CIFAR-10 classification task. For this purpose, we utilize a simple predictor, i.e., a four-layer MLP with ReLU nonlinearities, and jointly train the encoder and the predictor end-to-end in a supervised manner. We test for prediction as well as for zero shot prediction. Figure \ref{fig:Interpolation} displays the development of training loss against validation loss measured by means of the rooted-mean-squared error (RMSE). There are a few outliers in the NAS-Bench-101 graphs that end up with a low validation accuracy on the CIFAR-10 classification task. Figure \ref{fig:graphs} visualizes these outliers and shows that our model is able to find them even if it cannot perfectly predict their accuracies. One can see that the model predicts the ground truth of the other graphs very accurately. To further explore the loss, Figure \ref{fig:losses} illustrates the mean and variance of the squared error of the test set partitioned in 10 bins with respect to the ground truth accuracy. The greater part of the loss arises from graphs with a low accuracy. More importantly, our model is very accurate in its prediction for graphs of interest namely graphs with high accuracy. The rather bad prediction of graphs with low and intermediate accuracy can be explained through their low share in the dataset, see Figure \ref{fig:numbers} (left).

\begin{figure}
    \centering
    \begin{subfigure}[b]{0.23\textwidth}
        \includegraphics[width=\textwidth]{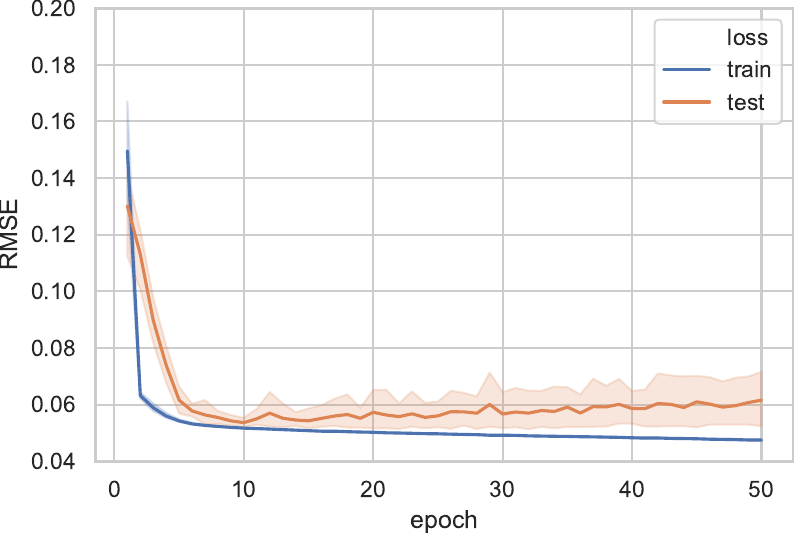}
    \end{subfigure}
    \begin{subfigure}[b]{0.23\textwidth}
        \includegraphics[width=\textwidth]{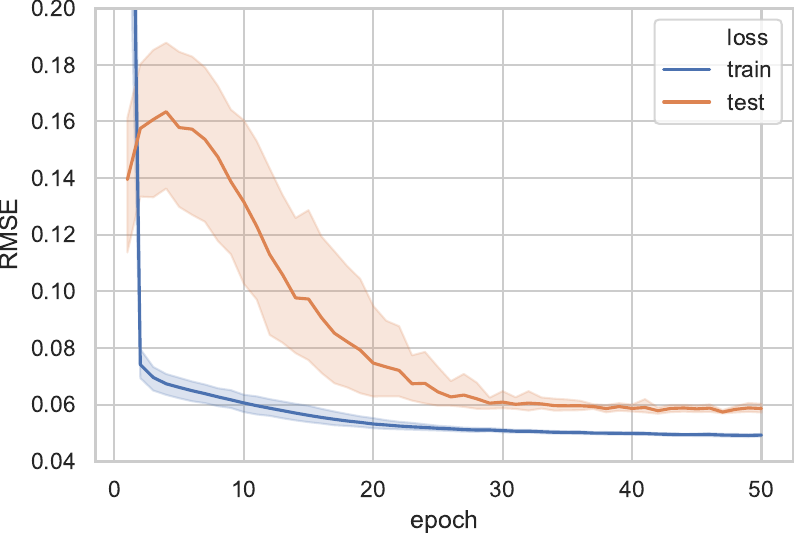}
    \end{subfigure}
    \caption{Progress of loss and test error over 50 epochs regarding zero shot prediction with two distinct splits. One training set consists of all graphs of length ${2,3,4,5,7}$ with a test set of the graphs of length $6$ (left). The other consists of all graphs of length ${2,3,4,5,6}$ with a test set of the graphs of length $7$ (right).}\label{fig:extrapolation}
\end{figure}

Next, we consider the task of predicting the validation accuracy of previously unseen graph types, i.e., zero-shot prediction.
The zero shot prediction task is furthermore divided into two subtasks. First, the encoder is learned on all graphs of length ${2,3,4,5,7}$ and tested on graphs of length $6$. Second, we learn the encoder on graphs of length ${2,3,4,5,6}$ and test it on graphs of length $7$. The progress of the loss and the test error can be seen in Figure \ref{fig:extrapolation}. The experiments show that our model is able to accurately predict data that it has never seen before. The behavior of the test error during the second zero shot prediction task, see Figure \ref{fig:extrapolation} (right), displays interesting information. During the first epochs, the error rises before it starts decreasing and approaching the training loss asymptotically. One interpretation could be that the model first learns simple graph properties like the number of nodes before it learns more complex graph substructures that generalize to the unseen data.

Table \ref{tab:rmse} summarizes the performance prediction results. All experiments are repeated $5$ times and we report the mean and the relative standard deviation.

\begin{table}[t]
	\begin{center}\begin{footnotesize}
			\begin{tabular}{l|c|c|c}
				\toprule

				& \multicolumn{1}{c|}{Prediction} & \multicolumn{2}{c}{Zero Shot Prediction} \\
				&   &  \footnotesize{$2,3,4,5,7 -6$} & \footnotesize{$2,3,4,5,6 -7$} \\
				\midrule
				VS-GAE & $0.0487(\pm 0.3\%)$ & $0.0526(\pm 4\%)$ & $0.0571(\pm 1.4\%)$ \\

				\bottomrule
			\end{tabular}
	\end{footnotesize}\end{center}
	\caption{Predictive performance of the VS-GAE encoder in terms of RMSE on prediction and zero shot prediction.}
	\label{tab:rmse}
\end{table}

 
\begin{table}[t]
\begin{center}\begin{footnotesize}
\begin{tabular}{l|cccc}
\toprule
 Method & Accuracy & Validity \\
\midrule
 VS-GAE  &   99.99 & 95.09  \\
\bottomrule
\end{tabular}
\end{footnotesize}\end{center}
\caption{The reconstruction accuracy and prior validity of VS-GAE in \%.}

\label{tab:VAE_Abilites}
\end{table}

\subsection{Model Ability} \label{sec:Model_Ability}
In this section, we evaluate VS-GAE by means of reconstruction ability and valid generation of neural architectures. To evaluate these abilities, we train the VS-GAE on $90 \%$ of the dataset and test it on the $10\%$ held-out data.

We first measure the reconstruction accuracy which describes how often our model can reconstruct the input graphs of the test set perfectly.  For this purpose, after calculating the mean $h_G$ and the variance $h^{\mathrm{var}}_G$ of the approximated posterior $q_{\phi}(\textbf{z} \vert G)$ for the test set, we sample $\textbf{z}$ from the latent representation of each input graph $10$ times and decode each sample again $10$ times. The average portion of the decoded graphs that are identical to the input ones is then reported as the reconstruction accuracy. 

The second ability we are interested in is the prior validity which quantifies how often our model is able to generate valid graphs from the VS-GAE prior distribution. Following \cite{zhang2019d}, we sample $1,000$ vectors from the latent space with prior distribution $p(\textbf{z})$ and decode each vector $10$~times. The average portion of the decoded graphs that are valid is then reported as the prior validity. For a valid graph by means of the NAS-Bench 101 \cite{ying2019bench} search space, it has to pass the following validity checks: 1) exactly one starting point, i.e., the input node, 2) exactly one ending point, i.e., the output node, 3) there exist no nodes which do not have any predecessors, except for the input node, 4)~there exist no nodes which do not have any successors, except for the output node, 5) the graphs are DAGs.

See Table \ref{tab:VAE_Abilites} for the evaluation results. Our model, VS-GAE, shows a nearly perfect reconstruction accuracy with a high prior validity. Thus, VS-GAE represents a strong tool for further downstream tasks like sampling or neural architecture generation.

\subsection{Stability regarding Sampling}

\begin{figure}
    \centering
    \begin{subfigure}[b]{0.23\textwidth}
        \includegraphics[width=\textwidth]{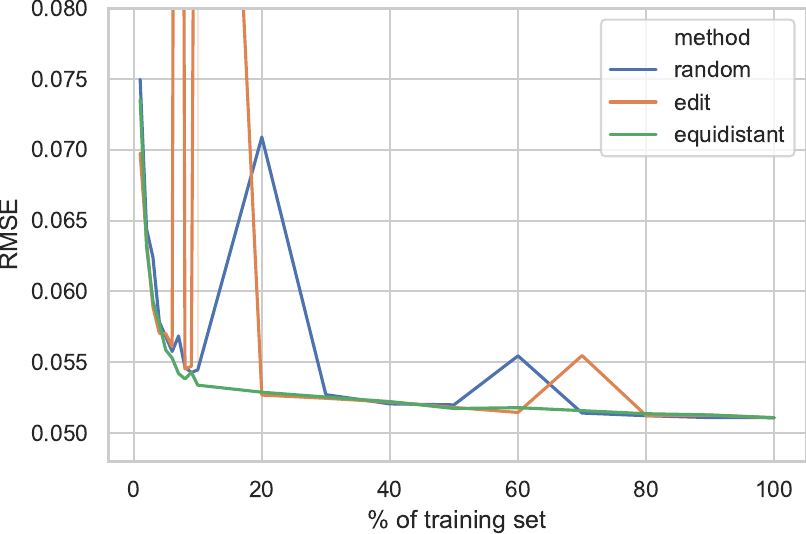}
    \end{subfigure}
    \begin{subfigure}[b]{0.23\textwidth}
        \includegraphics[width=\textwidth]{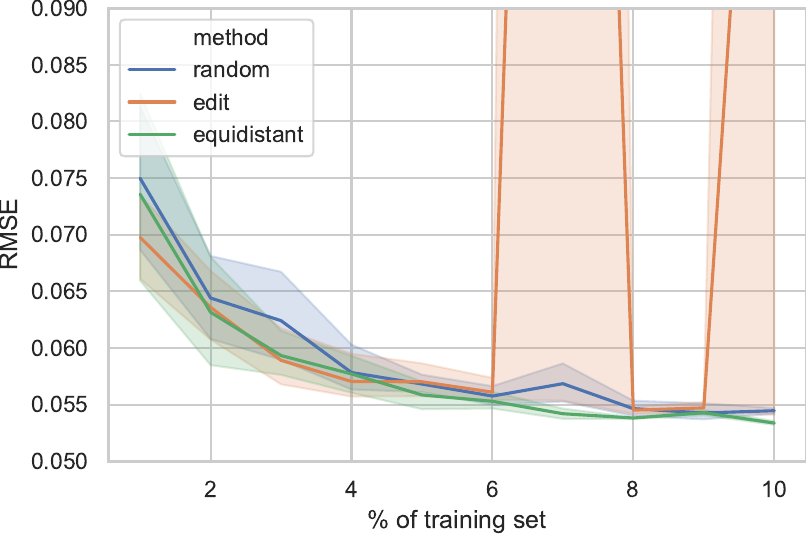}
    \end{subfigure}
    \caption{The test error regarding sample method and sample size visualized in two different resolutions. Sampling based on discrete methods can yield very inaccurate predictions while sampling in the VS-GAE space is most stable.}\label{fig:sampling}
\end{figure}

In this experiment we use the different sampling methods from Section \ref{sec:sampling} to downsize the training set with different factors ranging from $1\%$ to $90\%$ of the original size. We keep the validation set and the test set unchanged and calculate the test error regarding the best validation error over $100$ epochs. The results are outlined in Figure \ref{fig:sampling}. While the error fluctuates drastically for the sampling methods in the discrete space, the latent space created by VS-GAE enables a sampling that allows for stable accuracy predictions. Interestingly, we can not confirm that sampling uniformly by means of the edit distance gives any stability advantage over sampling uniformly at random.

\section{Conclusion}\label{sec:conclusion}
In this paper, we proposed VS-GAE, a new variational-sequential graph autoencoder, specialized on graphs representing neural architectures. Through multiple experiments on NAS-Bench-101, we examined various capabilities of VS-GAE. On the one hand, its encoder is a powerful tool regarding performance prediction, even in the zero-shot setup. On the other hand, the latent space generated by VS-GAE enables useful sampling methods in the graph space.
Further research will mainly review the possibilities of neural architecture generation of VS-GAE, potentially in accordance with further performance prediction.

\section*{Acknowledgement}
The authors acknowledge support by the German Federal Ministry of Education and Research Foundation via the project DeToL.

{\small
\bibliographystyle{ieee_fullname}
\bibliography{egbib}
}

\newpage
\clearpage
\pagenumbering{roman}

\appendix
 \begin{center}
\section*{Supplementary Material \\ A Variational-Sequential Graph Autoencoder for Neural Architecture Performance Prediction}  
\end{center}

\section{Model Implementation Details}
In this section, we give further details about the implementation of our proposed models.

\subsection{The Encoder}
\paragraph{Message}
The message module $M^{(t)}$ concatenates the embedding of the considered node $h_v^{(t-1)}$ as well as the incoming embedding $h_u^{(t-1)}$, each of dimension $d_n$. It further performs a linear transformation on the concatenated embedding. The reverse message module $M_{out}^{(t)}$ is a clone of $M^{(t)}$ initialized with its own weights,
\begin{align*}
M^{(t)} &= \mathrm{Lin}_{2d_n \times 2d_n}\bigl([h_v^{(t-1)}, h_u^{(t-1)}]\bigr),\\
M_{out}^{(t)} &= \mathrm{Lin'}_{2d_n \times 2d_n}\bigl([h_v^{(t-1)}, h_u^{(t-1)}]\bigr).
\end{align*}
The message module (green) and the reverse message (red) can be seen on the left side of Figure \ref{fig:encoder} in the main paper.

\paragraph{Update}
The update module $U^{(t)}$ is a single GRU cell. First, the incoming messages $m_{u\rightarrow v}$ and $m^{out}_{u\rightarrow v}$ are added and handled as the GRU input. Second, the node embedding $h_v^{(t-1)}$ is treated as the hidden state and is updated,
\begin{align*}
U^{(t)} = \mathrm{GRUCell}_{2d_n, d_n}\bigl(m_{u\rightarrow v} + m^{out}_{u\rightarrow v},\  h_v^{(t-1)}\bigr).
\end{align*}

\paragraph{Aggregation}
We use two rounds of propagation before aggregating the node embeddings into a single graph embedding. This graph aggregation consists of two parts. First, a linear layer transforms the node embeddings to the required graph embedding dimension $d_g$. Second, another linear layer combined with a sigmoid handles each node's fraction in the graph embedding,
\begin{align*}
A_1 &= \mathrm{Lin}_{d_n \times 2d_n}(h_v^{(2)}),\\
A_2 &= \sigma\bigl(\mathrm{Lin}_{d_n \times 1}(h_v^{(2)})\bigr),\\
A &= \sum_v A_1 \odot A_2.
\end{align*}
The aggregation function for the variational outputs $\tilde{A}$ is an exact copy of $A$ with its own weights. An illustration of the aggregation module is given in Figure \ref{fig:encoder} (right).

\subsection{VS-GAE}
\paragraph{InitNode}
The learnable embedding look-up table $E$ consists of five embeddings of size $d_n$, one for each of the five node types. It is initialized from $\mathcal{N}(0,1)$.
The InitNode module concatenates the sampled point of the latent space $\textbf{z}$ of size $d_g$, the summary of the partially created graph $h_{G^{(t)}}$ of size $d_g$ and the node embedding of the picked node type,
\begin{align*}
f^{1}_{\mathrm{initNode}} &= \mathrm{Lin}_{2d_g+d_n \times d_g+d_n}([\textbf{z}, h_{G^{(t)}}, L(\mathrm{type})]),\\
f_{\mathrm{initNode}} &= \mathrm{Lin}_{d_g+d_n \times d_n}\bigl(\mathrm{ReLU}(f^{1}_{\mathrm{initNode}})\bigr).
\end{align*}
This can be seen in Figure \ref{fig:decoder} c). For the very first embedding, we exclude the partially created graph,
\begin{align*}
f^{1}_{\mathrm{startNode}} &= \mathrm{Lin}_{d_g+d_n \times d_g+d_n}([\textbf{z}, L(\mathrm{type})]),\\
f_{\mathrm{startNode}} &= \mathrm{Lin}_{d_g+d_n \times d_n}\bigl(\mathrm{ReLU}(f^{1}_{\mathrm{startNode}})\bigr).
\end{align*}

\paragraph{GraphProp}
The GraphProp module consists of two rounds of message passing with the exact same modules from above and a variance-free graph aggregation. Each of these modules is initialized with its own weights, see Figure~\ref{fig:decoder}~a).

\paragraph{AddNode}
The AddNode module concatenates the sampled point of the latent space $\textbf{z}$ and the summary of the partially created graph $h_{G^{(t)}}$ and outputs logits over all five possible node types,
\begin{align*}
f^{1}_{\mathrm{addNode}} &= \mathrm{Lin}_{2d_g \times d_g}([\textbf{z}, h_{G^{(t)}}]),\\
f_{\mathrm{addNode}} &= \mathrm{Lin}_{d_g \times 5}\bigl(\mathrm{ReLU}(f^{1}_{\mathrm{addNode}})\bigr).
\end{align*}
This can be seen in Figure \ref{fig:decoder} b).

\paragraph{AddEdges}
The AddEdges module concatenates the embedding of the newly created node $h_{t+1}$ and each previous node $h_v$ as well as the sampled point of the latent space $\textbf{z}$ and the summary of the partially created graph $h_{G^{(t)}}$,
\begin{align*}
f^{1}_{\mathrm{addEdges}} &= \mathrm{Lin}_{2d_g+2d_n \times d_g+d_n}([h_{t+1}, h_v, \textbf{z}, h_{G^{(t)}}]),\\
f_{\mathrm{addEdges}} &= \mathrm{Lin}_{d_g+d_n \times 1}\bigl(\mathrm{ReLU}(f^{1}_{\mathrm{addEdges}})\bigr).
\end{align*}
The output is the score that describes the probability of an edge in logits. This process is illustrated in Figure \ref{fig:decoder} d).

\section{Experiment Details}
The node and graph dimensions, $d_n=250$ and $d_g=56$, are chosen as in \cite{Jin2018,zhang2019d} to attain comparability. For all experiments, we used the Adam optimizer with no dropouts. For training VS-GAE, we used a learning rate of $1e^{-4}$ for a total amount of $300$ epochs. Whenever the loss didn't drop for $10$ epochs, we decreased the learning rate by a factor of~$1e^{-1}$. 

\subsection{Performance Prediction}
The hidden layers of the predictor are of size $28, 14$ and $7$. We used no activation function for the very last output (linear regression) and trained the joint model with a learning rate of~$1e^{-5}$.

\end{document}